\documentclass{article}
\usepackage{spconf,amsmath,epsfig}
\usepackage{amsthm}
\usepackage{url}
\usepackage{multirow}
\usepackage{rotating}
\usepackage{amssymb}
\usepackage{graphicx,amsfonts}
\usepackage{algorithm,algorithmic}

\newcommand{\cH}{\mathcal{H}}

\newcommand{\cL}{\mathcal{L}}

\newcommand{\cC}{\mathcal{C}}

\newcommand{\bZ}{\boldsymbol{Z}}

\newcommand{\bR}{\boldsymbol{R}}
\newcommand{\bS}{\boldsymbol{S}}

\newcommand{\bT}{\boldsymbol{T}}

\newcommand{\bPhi}{\boldsymbol{\Phi}}

\newcommand{\bx}{\boldsymbol{x}}
\newcommand{\bh}{\boldsymbol{h}}
\newcommand{\by}{\boldsymbol{y}}

\newcommand{\bz}{\boldsymbol{z}}
\newcommand{\ba}{\boldsymbol{a}}
\newcommand{\bb}{\boldsymbol{b}}

\newcommand{\bc}{\boldsymbol{c}}

\newcommand{\bw}{\boldsymbol{w}}

\newcommand{\bbf}{\boldsymbol{f}}
\newcommand{\bg}{\boldsymbol{g}}
\newcommand{\bZero}{\boldsymbol{0}}

\newcommand{\beq}{\begin{equation}}
\newcommand{\eeq}{\end{equation}}
\newcommand{\beqn}{\begin{eqnarray}}
\newcommand{\eeqn}{\end{eqnarray}}
\newcommand{\beqns}{\begin{eqnarray*}}
\newcommand{\eeqns}{\end{eqnarray*}}

\newcommand{\R}{\mathbb{R}}
\newcommand{\HH}{\mathbb{H}}
\newcommand{\XX}{\mathbb{X}}
\newcommand{\C}{\mathbb{C}}

\newcommand{\N}{\mathbb{N}}

\newcommand{\frechet}{\textrm{Fr\'{e}chet }}
\newcommand{\fredif}{\textrm{Fr\'{e}chet differentiable }}

\makeatletter
\newcommand{\bdiv}{\mathop{\operator@font div}}
\makeatother

\makeatletter
\newcommand{\diag}{\mathop{\operator@font diag}}
\makeatother

\makeatletter
\newcommand{\conv}{\mathop{\operator@font conv}}
\makeatother

\makeatletter
\newcommand{\sign}{\mathop{\operator@font sign}}
\makeatother \makeatletter

\makeatletter
\newcommand{\proj}{\mathop{\operator@font proj}}
\makeatother \makeatletter

\makeatletter
\newcommand{\spa}{\mathop{\operator@font span}}
\makeatother \makeatletter

\makeatletter
\newcommand{\epi}{\mathop{\operator@font epi}}
\makeatother \makeatletter

\makeatletter
\newcommand{\dom}{\mathop{\operator@font dom}}
\makeatother \makeatletter

\theoremstyle{remark}

\theoremstyle{definition}

\theoremstyle{definition}

\title{Extension of Wirtinger Calculus in RKH Spaces and the Complex Kernel LMS}
\name{Pantelis Bouboulis, Sergios Theodoridis}
\address{\emph{Department of Informatics and Telecommunications,}\\
\emph{University of Athens,}\\
\emph{Athens, Greece.}\\
\emph{\{bouboulis,stheodor\}@di.uoa.gr}
}
\begin{document}
\ninept
\maketitle
\begin{abstract}\label{SEC:ABS}
Over the last decade, kernel methods for nonlinear processing have successfully been used  in the machine learning community. However, so far, the emphasis  has been on batch techniques. It is only recently, that online adaptive techniques have been considered in the context of signal processing tasks. To  the best of our knowledge, no kernel-based strategy has been developed, so far, that is able to deal with complex valued signals. In this paper, we take advantage of a technique called \textit{complexification} of real RKHSs to attack this problem. In order to derive gradients and subgradients of operators that need to be defined on the associated complex RKHSs, we employ the powerful tool of Wirtinger's Calculus, which has recently attracted much attention in the signal processing community. Writinger's calculus simplifies computations and offers an elegant tool for treating complex signals. To this end, in this paper, the notion of Writinger's calculus is extended, for the first time, to include complex RKHSs and use it to derive the Complex Kernel  Least-Mean-Square (CKLMS) algorithm. Experiments verify that the CKLMS can be used to derive nonlinear stable algorithms, which offer significant performance improvements over the traditional complex LMS or Widely Linear complex LMS (WL-LMS) algorithms, when dealing with nonlinearities.
\end{abstract}

\section{Introduction}\label{SEC:INTRO}
Processing in Reproducing Kernel Hilbert Spaces (RKHSs) in the context of online adaptive processing is gaining in popularity within the Signal Processing community \cite{LiPokPrin, KivSmoWil, EngManMe, SlaTheYam, SlaTheYam2, SlaThe}. The main advantage of  mobilizing the tool of RKHSs is that the original nonlinear task is ``transformed" into a linear one, where one can employ an easier ``algebra".  Moreover, different types of nonlinearities can be treated in a unifying way, that does not affect  the derivation of the algorithms, except at the final implementation stage. The main concepts of this procedure can be summarized in the following two steps: 1) Map the finite dimensionality input data from the input space $F$ (usually $F\subset \R^\nu$) into a higher dimensionality (possibly infinite) RKHS $\cH$ and 2) Perform a linear processing (e.g., adaptive filtering) on the mapped data in $\cH$. The procedure is equivalent with a non-linear processing (non-linear filtering) in $F$.

An alternative way of describing this process is through the popular \textit{kernel trick} \cite{SchoSmo, TheoKou}: Given an algorithm, which is formulated in terms of dot products, one can construct an alternative algorithm by replacing each one of the dot products with a positive definite kernel $\kappa$.  The specific choice of kernel implicitly defines  an RKHS with an appropriate inner product. Furthermore, the choice of kernel also defines the type of nonlinearity that underlies the model to be used. The main representatives of this class of algorithms are the celebrated \textit{support vector machines} (SVMs), which have dominated the research in machine learning over the last decade. Besides SVMs and the more recent applications in adaptive filtering, there is a plethora of other scientific domains that have gained from adopting kernel methods (e.g., image processing and denoising \cite{KimFraScho, BouSlaThe}, principal component analysis \cite{SchoSmoMu}, clustering \cite{FilCaMaRo}, e.t.c.).

In this paper, we focus on the recently developed \textit{Kernel Least Mean Squares Algorithm} (KLMS), which is the LMS algorithm in RKHSs \cite{LiPokPrin, LiuPriHay}. KLMS, as all the known kernel methods that use real-valued kernels, is able to deal with real valued data sequences only. To our knowledge, no kernel-based strategy has been developed, so far, that is able to effectively deal with complex valued signals. The main contributions of this paper are: a) the development of a wide framework that allows real-valued kernel algorithms to be extended to treat complex data effectively, taking advantage of a technique called \textit{complexification} of real RKHSs, b) the extension of \textit{Wirtinger's Calculus} in complex RKHSs as a means for the elegant and efficient computation of the gradients, that are involved in many adaptive filtering algorithms, and c) the development of the Complex Kernel LMS (CKLMS) algorithm, by exploiting the developed Wirtinger's calculus.  Wirtinger's calculus \cite{Wirti} is enjoying  increasing popularity, recently, mainly in the context of \textit{Widely Linear} complex adaptive filters \cite{Picin95, ManGoh, Adali10, Adali08a, Adali08b, MatPaSte, CaGePaVe, Moreno}, providing a tool for the derivation of gradients in the complex domain.

The paper is organized as follows. We start with a minimal introduction to RKHSs in Section \ref{SEC:PRELIM}, before we briefly review the KLMS algorithm in Section \ref{SEC:KLMS}. In Section \ref{SEC:Complex}, we describe the complexification procedure of a real RKHS that provides the main framework for complex kernel methods, based on the popular real valued reproducing kernels (e.g., gaussian, polynomial, e.t.c.). The main notions of the extended Wirtinger's Calculus are summarized in Section \ref{SEC:Wirtinger} and the CKLMS is developed thereafter in Section \ref{SEC:CKLMS}. Finally, experimental results and conclusions are provided in Sections \ref{SEC:Experim} and \ref{SEC:Concl} respectively. We will denote the set of all integers, real and complex numbers by $\N$, $\R$ and $\C$ respectively. Vector or matrix valued quantities appear in boldfaced symbols.

\section{Reproducing Kernel Hilbert Spaces}\label{SEC:PRELIM}
We start with some basic definitions regarding RKHSs. Let $X$ be a non empty set with $\bx_1,\dots,\bx_N \in X$.
Consider a Hilbert space $\cH$ of real valued functions, $f$, defined on a set $X$, with a corresponding inner product $\langle\cdot,\cdot\rangle_\cH$. We will call $\cH$ as a \textit{Reproducing Kernel Hilbert Space} - RKHS, if there exists a function, known as kernel, $\kappa:X\times X\rightarrow\R$ with the following two properties:
\begin{enumerate}
\item For every $\bx\in X$, $\kappa(\bx,\cdot)$ belongs to $\cH$.
\item $\kappa$ has the so called \textit{reproducing property}, i.e.
$f(\bx)=\langle f,\kappa(\bx,\cdot)\rangle_\cH, \textrm{ for all } f\in\cH$.
In particular:
$$\kappa(\bx,\by)=\langle \kappa(\bx,\cdot),\kappa(\by,\cdot)\rangle_\cH.$$
\end{enumerate}

In can been shown that the kernel $\kappa$ generates the entire space $\cH$, i.e. $\cH=\overline{\spa\{\kappa(\bx,\cdot)|\bx\in X\}}.$
There are several kernels that are used in practice (see \cite{SchoSmo}). Among the most widely used are the the polynomial kernel: $\kappa(\bx,\by) = \left(1 + x^T y\right)^d$, $d\in\N$ and the gaussian kernel: $\kappa(\bx,\by)=exp\left(-\|\bx-\by\|^2/\sigma^2\right)$, $\sigma>0$.

Although there exist complex reproducing kernels that give rise to RKHSs of complex valued functions \cite{Paulsen}, in this paper we focus our attention on complexifying real valued ones, which have been extensively studied and contain several popular examples. Later on (in section \ref{SEC:Complex}), we will show how one can construct complex RKHSs from real ones, through a technique called complexification.

\section{Kernel LMS}\label{SEC:KLMS}
In a typical LMS filter the goal is to learn a linear input output mapping $f:X\rightarrow\R:f(\bx)=\bw^T\bx$, $X\subset\R^\nu$, based on a sequence of examples $(\bx(1),d(1)), (\bx(2),d(2)), \dots, (\bx(N),d(N))$, so that to minimize the mean square error, $E\left[|d(n) - \bw^T\bx(n)|^2\right]$. To this end, the gradient descent rationale is employed and at each time instant,  $n=1,2,\dots,N$, the gradient of $E[e(n)\bx(n)]$ is estimated via its current measurement, i.e.,  $\hat E[e(n)\bx(n)]=e(n)\bx(n)$, where $e(n) = d(n) - \bw(n-1)^T\bx(n)$ is the error at instance $n=2,\dots,N$. It takes a few lines of elementary algebra to deduce that the update of the unknown vector parameter is: $\bw(n) = \bw(n-1) + \mu e(n) \bx(n)$, where $\mu$ is the step update. If we take the initial value of $\bw$ as $\bw(0)=\bZero$, then the repeated application of the update equation yields:
\begin{align}\label{EQ:LMS1}
\bw(n) = \mu \sum_{k=1}^n e(k) \bx(k)
\end{align}
Hence, for the filter output at instance $n$ we have:
\begin{align}\label{EQ:LMS2}
\hat d(n) = \bw(n-1)^T\bx(n) = \mu \sum_{k=1}^{n-1} e(k) \bx(k)^T \bx(n),
\end{align}
for $n=1,2,\dots,N$. Equation (\ref{EQ:LMS2}) is expressed in terms of inner products only, hence it allows for the application of the kernel trick. Thus, the filter output of the KLMS at instance $n$ is
\begin{align}\label{EQ:KLMS2}
\hat d(n) =  \left\langle \bx(n), \bw(n-1)\right\rangle = \mu\sum_{k=1}^{n-1} e(k) \kappa\left( \bx(n), \bx(k)\right),
\end{align}
\begin{align}\label{EQ:KLMS1}
\textrm{while }\hspace{4em} \bw(n) = \mu \sum_{k=1}^n e(k) \kappa(\bx(k),\cdot),
\end{align}
for $n=1,2,\dots,N$.

Another, more formal way of developing the KLMS is the following. First, we transform the input space $X$ to a high dimensional feature space $\cH$ through the (implicit) mapping $\Phi:X\rightarrow\cH$, $\Phi(\bx)=\kappa(\bx,\cdot)$. Thus, the training examples become
$$(\Phi(\bx(1)), d(1)), \dots, (\Phi(\bx(N)), d(N)).$$
We apply the LMS procedure on the transformed data, with the linear filter output $\hat d(n)=\langle \Phi(\bx(n)), \bw\rangle$. The model $\langle \Phi(\bx), \bw\rangle$ is more representive than the simple $\bw^T\bx$, since it includes the nonlinear modeling through the presence of the kernel. The objective now becomes to minimize the cost function
\begin{align*}
E\left[|d(n) - \langle \Phi(\bx(n)), \bw\rangle|^2\right].
\end{align*}
Using the notion of the \frechet derivative, which has to be mobilized, since the dimensionality of the RKHS may be infinite, we are able to derive the gradient of the aforementioned cost function with respect to $\bw$.  It has to be emphasized, that now $\bw$ is not a vector, but a function, i.e., a point in the linear Hilbert space. It turns out that the update of the KLMS is given by $\bw(n)=\bw(n-1) + \mu e(n) \Phi(\bx(n))$, where $e(n)=d(n)-\hat d(n)$. From this update, following the same procedure as in LMS and applying the reproducing property, we obtain equations (\ref{EQ:KLMS2}) and (\ref{EQ:KLMS1}), which are at the core of the KLMS algorithm. More details and the algorithmic implementation may be found in \cite{LiuPriHay}.

Note that, in a number of attempts to kernelize known algorithms, that are cast in inner products, the kernel trick is, usually, used in a "black box" rationale, without consideration of the problem in the RKH space, in which the (implicit) processing is carried out. Such an approach, often, does not allow for a deeper understanding of the problem, especially if a further theoretical analysis is required. Moreover, in our case, such a "blind" application of the kernel trick on a standard complex LMS form, can only lead to spaces defined by complex kernels. Complex RKH spaces, built around complexification of real  kernels, do not result as a direct application of the standard kernel trick.

\section{Complexification of a real RKHS}\label{SEC:Complex}
To generalize the kernel adaptive filtering algorithms on complex domains, we need a generalized framework regarding complex RKHSs. In this paper, we employ a simple technique called \textit{complexification} of real RKHSs, which has the advantage of allowing  modeling in complex RKHSs using popular well-established real kernels (e.g., gaussian, polynomial, e.t.c.).

Let $X\subseteq\R^\nu$. Define $X^2=X\times X\subseteq\R^{2\nu}$ and $\XX=\{\bx+i\by, \bx,\by\in X\}$ equipped with a complex product structure. Let $\cH$ be a real RKHS associated with a real kernel $\kappa$ defined on $X^2\times X^2$ and let $\langle\cdot,\cdot\rangle_\cH$ be its corresponding inner product. Then, every $f\in\cH$ can be regarded as a function defined on either $X^2$ or $\XX$, i.e., $f(\bz) = f(\bx+i\by) = f(\bx,\by)$.

Next, we define $\cH^2=\cH\times\cH$. It is easy to verify that $\cH^2$ is also a Hilbert Space with inner product
\begin{align}
\langle \bbf, \bg\rangle_{\cH^2} = \langle f_1, g_1\rangle_\cH + \langle f_2, g_2\rangle_\cH,
\end{align}
for $\bbf=(f_1,f_2)^T$, $\bg=(g_1,g_2)^T$. Our objective is to enrich $\cH^2$ with a complex structure. We address this problem using the complexification of the real RKHS  $\cH$. To this end, we define the space $\HH= \{\bbf=f_1 + i f_2;\;f_1,f_2\in\cH\}$
equipped with the complex inner product:
\begin{align*}
\langle \bbf, \bg\rangle_{\HH}= &\langle f_1, g_1\rangle_\cH + \langle f_2, g_2\rangle_\cH + \\
                 &i\left(\langle f_2, g_1\rangle_\cH - \langle f_1, g_2\rangle_\cH\right),
\end{align*}
for $\bbf=f_1 + if_2$, $\bg=g_1 + i g_2$. It is not difficult to verify that $\HH$ is a complex RKHS with kernel $\kappa$ \cite{Paulsen}. We call $\HH$ the complexification of $\cH$.

To complete the presentation of the required framework for working on complex RKHSs, we need a technique to map the samples data from the complex input space to the complexified RKHS $\HH$. This problem will be addressed in section \ref{SEC:CKLMS}.

\section{Wirtinger's Calculus in complex RKHS}\label{SEC:Wirtinger}
Wirtinger's calculus \cite{Wirti} has become very popular in the signal processing community mainly in the context of complex adaptive filtering, as a means of computing, in an elegant way,  gradients of real valued cost functions defined on complex domains ($\C^\nu$). Such functions, obviously, are not holomorphic and therefore the complex derivative cannot be used. Instead, if we consider that the cost function is defined on a Euclidean domain with a double dimensionality ($\R^{2\nu}$), then the real derivatives may be employed. The price of this approach is that the computations become cumbersome and tedious. Wirtinger's calculus provides an alternative equivalent formulation, that is based on simple rules and principles and which bear a great resemblance to the rules of the standard complex derivative.

In the case of a simple non-holomorphic complex function $T$ defined on $U\subseteq\C$, Wirtinger's calculus considers two forms of derivatives, the \textit{$\R$-derivative} and the \textit{conjugate $\R$-derivative}, which are defined as follows:
\begin{align*}
\frac{\partial T}{\partial z}=\frac{1}{2}\left(\frac{\partial u}{\partial x}+\frac{\partial v}{\partial y}\right) + \frac{i}{2}\left(\frac{\partial v}{\partial x}-\frac{\partial u}{\partial y}\right),\\
\frac{\partial T}{\partial z^*}=\frac{1}{2}\left(\frac{\partial u}{\partial x}-\frac{\partial v}{\partial y}\right) + \frac{i}{2}\left(\frac{\partial v}{\partial x}+\frac{\partial u}{\partial y}\right)
\end{align*}
where $T(z)=T(x+iy)=T(x,y)=u(x,y)+i v(x,y)$. Note that any such non-holomorphic function can be written in the form $T(z,z^*)$, so that for fixed $z^*$, $T$ is $z$-holomorphic and for fixed $z$, $T$ is $z^*$-holomorphic \cite{Delga} (assuming of course that $T(x,y)$ has partial derivatives of any order). This fact underlies the development of Wirtinger's calculus. Having this in mind, $\frac{\partial T}{\partial z}$,  can be easily evaluated as the standard complex partial derivative taken with respect to $z$ (thus treating $z^*$ as a constant). Consequently,  $\frac{\partial T}{\partial z^*}$ is evaluated as the standard complex partial derivative taken with respect to $z^*$ (thus treating $z$ as a constant). For example, if $T(z,z^*)=z(z^*)^2$, then
\begin{align*}
\frac{\partial T}{\partial z} = (z^*)^2,\quad \frac{\partial T}{\partial z^*}=2zz^*.
\end{align*}
Similar principles and rules hold for a function of many complex variables (i.e., $U\subseteq\C^{\nu}$) \cite{Delga}.

Wirtinger's calculus has been developed only for operators defined on finite dimensional spaces, $\C^{\nu}$. Hence, this calculus cannot be used in RKH spaces, where the dimensionality of the function space can be infinite. To this end, Wirtinger's calculus needs to be generalized to a general Hilbert space, and this is one of the main contributions of the current paper. A rigorous presentation of this extension is out of the scope of the paper (due to lack of space). Nevertheless, we will present the main ideas and results. At the heart of the generalization lies the notion of the \frechet differentiability. Consider a Hilbert space $H$ over the field $F$  (typically $\R$ or $\C$). The operator $T:H\rightarrow F$ is said to be \textit{\fredif} at $f_0$, if there exists a $u\in H$, such that
\begin{align}\label{EQ:frechet2}
\lim_{\|h\|_{H}\rightarrow 0}\frac{T(f_0+h)-T(f_0)-\langle u, h\rangle_{H}}{\|h\|_{H}}=0,
\end{align}
where $\langle\cdot, \cdot\rangle_{H}$ is the dot product of the Hilbert space $H$ and $\|\cdot\|_H=\sqrt{\langle\cdot, \cdot\rangle_H}$ is the
induced norm. The element $u$ is usually called the gradient of $T$ at $f_0$.

Since our study involves mainly RKHS, we will present the necessary tools in that context. The generalization to a general Hilbert space has also been developed and follows  a similar path. Consider the spaces $\HH$ and $\cH^2$ defined in section \ref{SEC:Complex}. Let $\bT:\HH \rightarrow \C$, $\bT=T_1 + iT_2$ be the operator we seek to differentiate. Assume that $\bT=(T_1,T_2)^T$, $\bT(\bbf) = \bT(f_1+i f_2) = \bT(f_1,f_2) = T_1(f_1,f_2) + i T_2(f_1,f_2)$, is differentiable as an operator defined on $\cH^2$ and let $\nabla_{1}T_1$, $\nabla_{2}T_1$, $\nabla_{1}T_2$ and $\nabla_{2}T_2$ be the partial derivatives, with respect to the first ($f_1$) and the second ($f_2$) variable respectively. It turns out, proof is omitted due to lack of space, that if $\bT(f_1,f_2)$ has derivatives of any order, then it can be written in
the form $\bT(\bbf, \bbf^*)$, where $\bbf^*=f_1-i f_2$, so that for fixed $\bbf^*$, $\bT$ is $\bbf$-holomorphic and for
fixed $\bbf$, $T$ is $\bbf^*$-holomorphic. We may define the $\R$-derivative and the conjugate $\R$-derivative of $\bT$ as follows:
\begin{align}
\nabla_{\bbf}\bT &= \frac{1}{2}\left(\nabla_1 T_1 + \nabla_2 T_2\right) + \frac{i}{2}\left(\nabla_1 T_2 - \nabla_2 T_1\right)\\
\nabla_{\bbf^*}\bT &= \frac{1}{2}\left(\nabla_1 T_1 - \nabla_2 T_2\right) + \frac{i}{2}\left(\nabla_1 T_2 + \nabla_2 T_1\right).
\end{align}
The following properties can be proved (among others):
\begin{enumerate}
\item if $\bT$ is $\bbf$-holomorphic (i.e., it has a Taylor series expansion with respect to $\bbf$), then $\nabla_{\bbf^*}\bT=\bZero$.
\item if $\bT$ is $\bbf^*$-holomorphic (i.e., it has a Taylor series expansion with respect to $\bbf^*$), then $\nabla_{\bbf}\bT=\bZero$.
\item $\left(\nabla_{\bbf} \bT\right)^* = \nabla_{\bbf^*} \bT^*$.
\item $\left(\nabla_{\bbf^*} \bT\right)^* = \nabla_{\bbf} \bT^*$.
\item If $\bT$ is real valued, then $\left(\nabla_{\bbf} \bT\right)^* = \nabla_{\bbf^*} \bT$.
\item The first order Taylor expansion around $\bbf\in\HH$ is given by
\begin{align*}
\bT(\bbf+\bh) =& \bT(\bbf) + \langle \bh, \left(\nabla_{\bbf} \bT(\bbf)\right)^* \rangle_\HH \\
&+ \langle \bh^*, \left(\nabla_{\bbf^*} \bT(\bbf)\right)^* \rangle_\HH.
\end{align*}
\item If $\bT(\bbf)=\langle \bbf, \bw\rangle_\HH$, then $\nabla_{\bbf}\bT=\bw^*$, $\nabla_{\bbf^*}\bT=\bZero$.
\item If $\bT(\bbf)=\langle \bw, \bbf\rangle_\HH$, then $\nabla_{\bbf}\bT=\bZero$, $\nabla_{\bbf^*}\bT=\bw$.
\item If $\bT(\bbf)=\langle \bbf^*, \bw\rangle_\HH$, then $\nabla_{\bbf}\bT=\bZero$, $\nabla_{\bbf^*}\bT=\bw^*$.
\item If $\bT(\bbf)=\langle \bw, \bbf^*\rangle_\HH$, then $\nabla_{\bbf}\bT=\bw$, $\nabla_{\bbf^*}\bT=\bZero$.
\item If $\bR,\bS:\HH\rightarrow\C$ are $\bbf$-analytic and $\bT=\bR\cdot\bS$ then:
\begin{align*}
\nabla_{\bbf}\bT = \nabla_{\bbf}\bR\cdot\bS + \nabla_{\bbf}\bS\cdot\bR.
\end{align*}
\end{enumerate}

An important consequence of the above properties is that if $\bT$ is a real valued operator defined on $\HH$, then its first order Taylor's expansion is given by:
\begin{align*}
\bT(\bbf+\bh) & =  \bT(\bbf) + \langle \bh, \left(\nabla_{\bbf}\bT(\bbf)\right)^*\rangle_\HH + \langle \bh^*, \left(\nabla_{\bbf^*}\bT(\bbf)\right)^* \rangle_\HH\\
& = \bT(\bbf) + \langle \bh, \nabla_{\bbf^*}\bT(\bbf)\rangle_\HH + \left(\langle \bh, \nabla_{\bbf^*}\bT(\bbf)\rangle_\HH\right)^*\\
& = \bT(\bbf) + 2\cdot \Re\left[ \langle \bh, \nabla_{\bbf^*}\bT(\bbf)\rangle_\HH\right].
\end{align*}
However, in view of the Cauchy Riemann inequality we have:
\begin{align*}
\Re\left[ \langle \bh, \nabla_{\bbf^*}\bT(\bbf)\rangle_\HH\right] & \leq \left|\langle \bh, \nabla_{\bbf^*}\bT(\bbf)\rangle_\HH\right|\\
&\leq \|\bh\|_\HH \cdot \| \nabla_{\bbf^*}\bT(\bbf)\|_\HH.
\end{align*}
The equality in the above relationship holds if $h\propto \nabla_{\bbf^*}\bT$. Hence, the direction of increase of $\bT$ is $\nabla_{\bbf^*}\bT(\bbf)$. Therefore, any gradient descent based algorithm minimizing $\bT(\bbf)$ is based on the update scheme:
\begin{align}
\bbf_{n} = \bbf_{n-1} - \mu\cdot\nabla_{\bbf^*}\bT(\bbf_{n-1}).
\end{align}

\section{Complex Kernel LMS}\label{SEC:CKLMS}
Consider the sequence of examples $(\bz(1),d(1))$, $(\bz(2),d(2))$, $\dots$, $(\bz(N),d(N))$, where $d(n)\in\C$, $\bz(n)\in V\subset\C^\nu$, $\bz(n)=\bx(n) + i \by(n)$, $\bx(n), \by(n)\in\R^\nu$, for $n=1,\dots,N$. We map the points $\bz(n)$ to the RKHS $\HH$ using the mapping $\bPhi$:
\begin{align*}
\bPhi(\bz(n)) &= \Phi(\bz(n)) + i \Phi(\bz(n)) \\
&= \kappa\left((\bx(n),\by(n))^T,\cdot\right) + i\cdot\kappa\left((\bx(n),\by(n))^T,\cdot\right),
\end{align*}
for $n=1,\dots,N$. The objective of the complex Kernel LMS is to minimize $E\left[\cL_n(\bw)\right]$, where
\begin{align*}
\cL_n(\bw) &=  |e(n)|^2 = |d(n) - \langle \bPhi(\bz(n)), \bw\rangle_\HH|^2\\
&= \left(d(n) - \langle \bPhi(\bz(n)), \bw\rangle_\HH\right) \left(d(n) - \langle \bPhi(\bz(n)), \bw\rangle_\HH\right)^*\\
&= \left(d(n) - \langle \bw^*, \bPhi(\bz(n))\rangle_\HH\right) \left(d(n)^* - \langle \bw, \bPhi(\bz(n))\rangle_\HH\right),
\end{align*}
at each instance $n$. We then apply the complex LMS to the transformed data, using the rules of Wirtinger's calculus to compute the gradient $\nabla_{\bw^*}\cL_n(\bw)=-e(n)^*\cdot\bPhi(\bz(n))$.
Therefore the CKLMS update rule becomes:
\begin{align}
\bw(n) = \bw(n-1) + \mu e(n)^*\cdot\bPhi(\bz(n)),
\end{align}
where $\bw(n)$ denotes the estimate at iteration $n$.

Assuming that $\bw(0)=\bZero$, the repeated application of the weight-update equation gives:
\begin{align}
\bw(n) = & \bw(n-1) + \mu e(n)^*\bPhi(\bz(n))\nonumber\\
= & \bw(n-2) + \mu e(n-1)^*\bPhi(\bz(n-1))\nonumber \\
& + \mu e(n)^*\bPhi(\bz(n))\nonumber\\
= & \sum_{k=1}^{n} e(k)^*\bPhi(\bz(k))\label{EQ:CKLMS_W}.
\end{align}
Thus, the filter output at iteration $n$ becomes:
\begin{align}
\hat d(n) =&\langle \bPhi(\bz(n)), \bw(n-1) \rangle_\HH \nonumber\\
=& \mu \sum_{k=1}^{n-1} e(k) \langle \bPhi(\bz(n)), \bPhi(\bz(k)) \rangle_\HH\nonumber\\
=& 2\mu \sum_{k=1}^{n-1} e(k) \kappa(\bz(n), \bz(k))\nonumber\\
=& 2\mu \sum_{k=1}^{n-1} \Re[e(n)]\kappa(\bz(n), \bz(k))\nonumber \\
&+ 2\mu \cdot i \sum_{k=1}^{n-1} \Im[e(n)]\kappa(\bz(n), \bz(k)),
\end{align}
where the evaluation of the kernel is done by replacing the complex vectors $\bz(n)$, of $\C^\nu$ with the corresponding real vectors of $\R^{2\nu}$, i.e.,
\begin{align*}
\bz(n)=\bx(n) + i\by(n) = (\bx(n),\by(n))^T.
\end{align*}

It can readily be shown that, since the CKLMS is the complex LMS in RKHS, the important properties of the LMS (convergence in the mean, misadjustment, e.t.c.) carry over to CKLMS. Furthermore, we may also define a normalized version, which we call \textit{Normalized Complex Kernel LMS} (NCKLMS). The weight-update of the NCKLMS is given by:
\begin{align*}
\bw(n) = & \bw(n-1) + \frac{\mu}{2\cdot\kappa(\bz(n),\bz(n))} e(n)^*\bPhi(\bz(n))
\end{align*}
The NCKLMS algorithm is summarized in Algorithm \ref{ALG:NCKLMS1}.

\begin{algorithm}[h!]
\caption{Normalized Complex Kernel  LMS}\label{ALG:NCKLMS1}
\textbf{INPUT: } $(\bz(1),d(1))$, $\dots$, $(\bz(N),d(N))$\\
\textbf{OUTPUT:} The expansion \\
$\bw=\sum_{k=1}^{N} a(k)\kappa(\bz(k),\cdot) + i\cdot \sum_{k=1}^{N} b(k)\kappa(\bz(k),\cdot)$.\\
\\
\textbf{Initialization:} Set  $\ba=\{\}$, $\bb=\{\}$, $\bZ=\{\}$ (i.e., $\bw=\bZero$). Select the step parameter $\mu$ and the kernel $\kappa$.
\begin{algorithmic}
\FOR{n=1:N}
\STATE{Compute the filter output:
\begin{align*}
\hat d(n) =& \sum_{k=1}^{n-1}(a(k)+b(k))\cdot\kappa(\bz(n),\bz(k))\\
     &+ \sum_{k=1}^{n-1}(a(k)-b(k))\cdot\kappa(\bz(n),\bz(k)).
\end{align*}}
\STATE{Compute the error: $e(n)=d(n)-\hat d(n)$.}
\STATE{$\gamma=2\kappa(\bz(n),\bz(n))$.}
\STATE{$a(n)=\mu(\Re[e(n)] + \Im[e(n)])/\gamma$.}
\STATE{$b(n)=\mu(\Re[e(n)] - \Im[e(n)])/\gamma$.}
\STATE{Add the new center $\bz(n)$ to the list of centers, i.e., add $\bz(n)$ to the list $\bZ$, add $a(n)$ to the list $\ba$, add $b(n)$ to the list $\bb$.}
\ENDFOR
\end{algorithmic}
\end{algorithm}

\begin{figure}
\begin{center}
\includegraphics[scale=0.45]{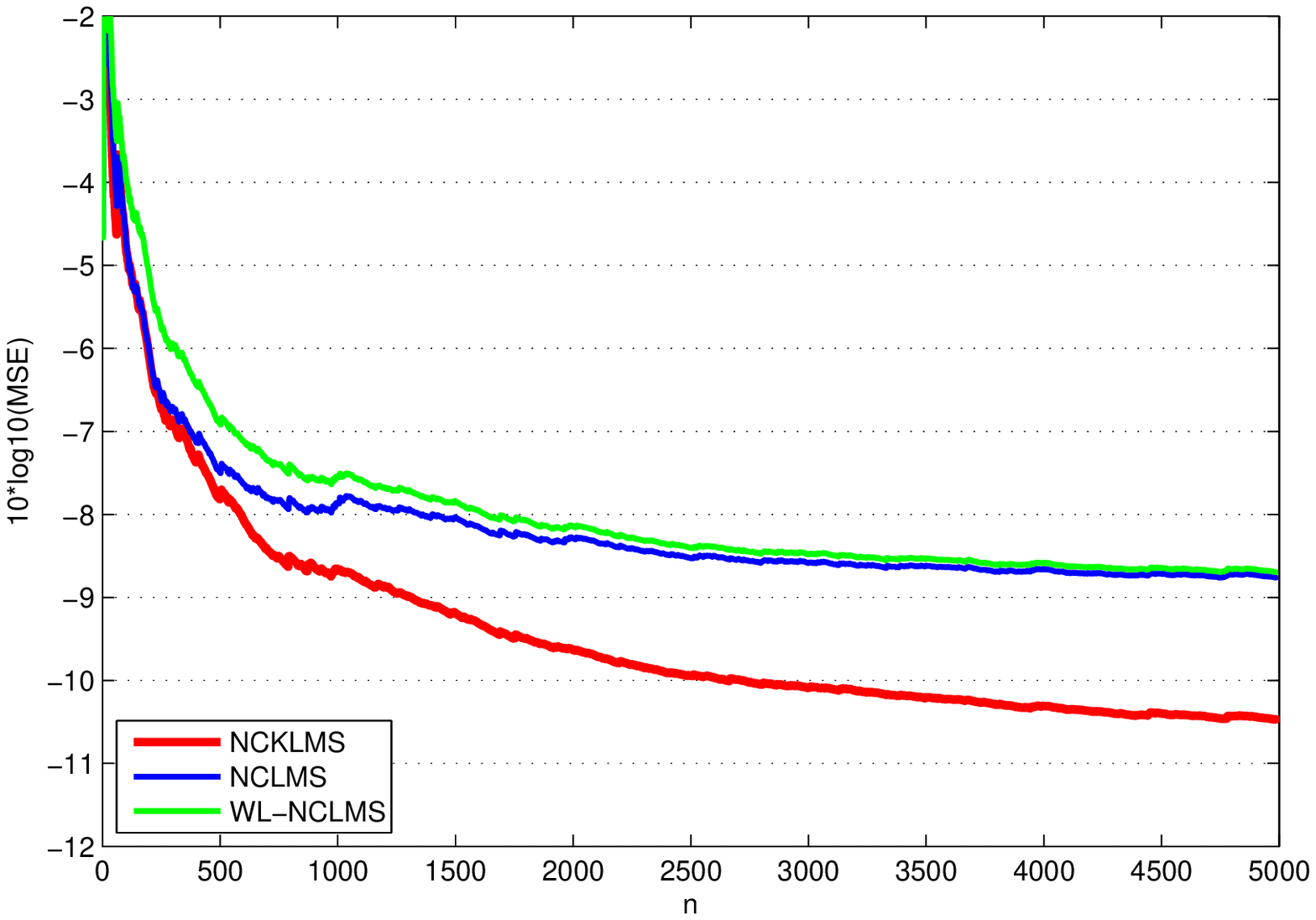}
\end{center}
\caption{Learning curves for KÍCLMS, ($\mu=1/2$) ÍCLMS ($\mu=1/16$) and WL-ÍCLMS ($\mu=1/16$) (filter length $L=5$, delay $D=2$) in the nonlinear channel equalization, for the circular input case.}\label{FIG:equal_circu}
\end{figure}

\subsection{Sparsification}
The main drawback of kernel based adaptive filtering algorithms is that they require a growing network of training centers $\bz_n$. They start from an empty set (usually called the \textit{dictionary}) and gradually add new samples to that set, to form a summation similar to the one shown in equation (\ref{EQ:CKLMS_W}). This results to an increasing memory and computational requirements, as time evolves. Several strategies have been proposed to cope with this problem and to produce sparse solutions. In this paper, we employ the well known \textit{novelty criterion} \cite{Platt, LiuPriHay}. In novelty criterion online sparsification, whenever a new data pair $(\bPhi(\bz_n),d_n)$ is considered, a decision is immediately made of whether to add the new center $\bPhi(\bz_n)$ to the dictionary of centers $\cC$. The decision is reached following two simple rules. First, the distance of the new center $\bPhi(\bz_n)$ from the current dictionary is evaluated: $dis = \min_{\bc_k\in\cC}\{\|\bPhi(\bz_n) - \bc_k\|_\HH\}$.
If this distance is smaller than a given threshold $\delta_1$ (i.e., the new center is close to the existing dictionary), then the center is not added to $\cC$. Otherwise, we compute the prediction error $e_n = d_n - \hat d_n$. If $|e_n|$ is smaller than a predefined threshold $\delta_2$, then the new center is discarded. Only if $|e_n|\geq\delta_2$ the new center $\bPhi(\bz_n)$ is added to the dictionary.

Besides the previous scenario, other scenarios are also possible, that keep the number updated parameters, per recursion, fixed. For example, the sliding window LMS can be used. In \cite{SlaTheYam, SlaTheYam2, SlaThe}, regularization, in the form of projections, has been used to cope efficiently with the problem. Results under such scenarios are available and will be presented elsewhere.

\section{Experiments}\label{SEC:Experim}
\begin{figure}
\begin{center}
\includegraphics[scale=0.35]{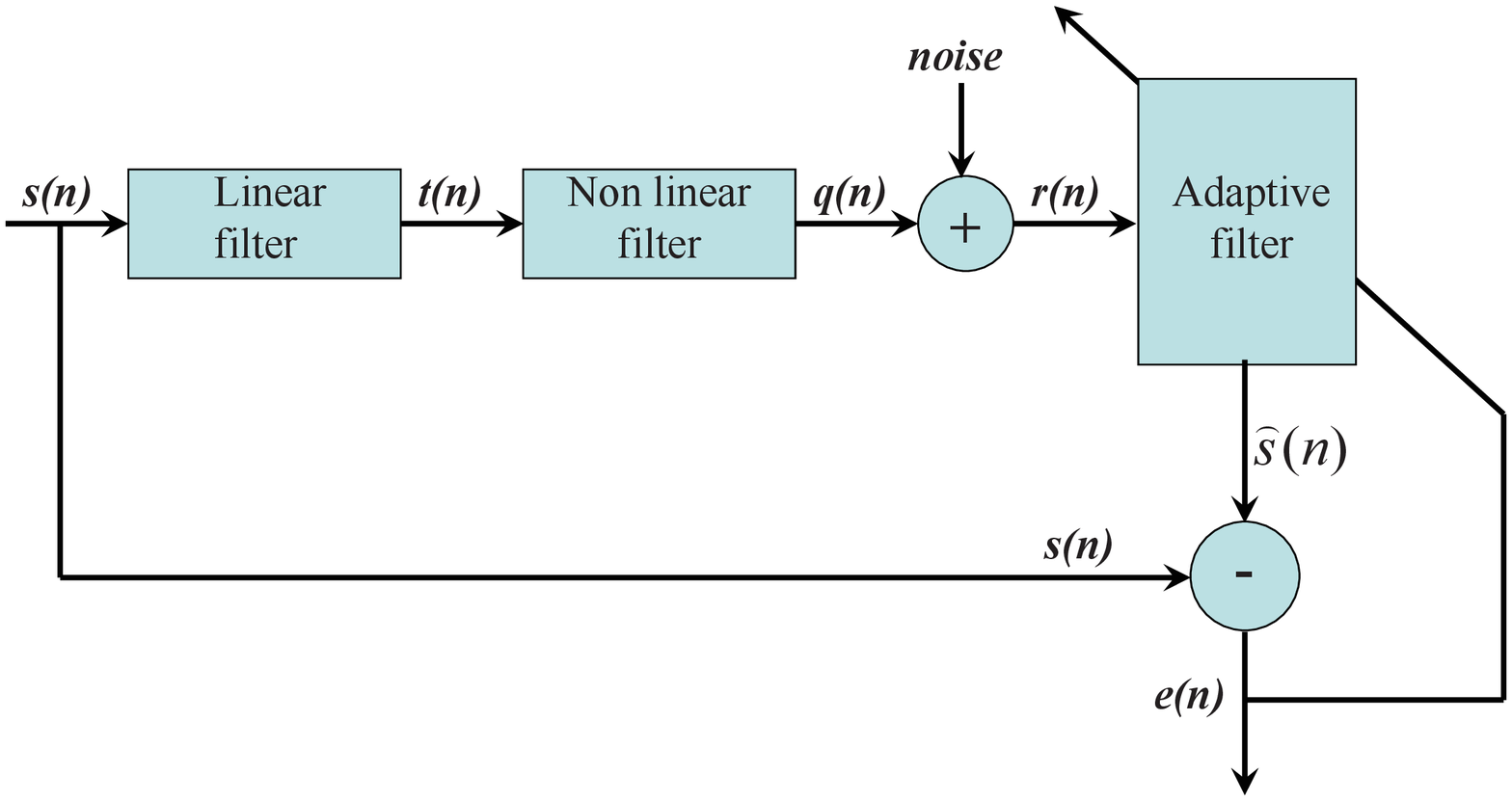}
\end{center}
\caption{The equalization problem.}\label{FIG:equal_form}
\end{figure}

We tested the CKLMS on a nonlinear channel equalization problem (see figure \ref{FIG:equal_form}). The nonlinear channel consists of a linear filter:
\begin{align*}
t(n)= (-0.9+0.8i)\cdot s(n) + (0.6-0.7i)\cdot s(n-1)
\end{align*}
and a memoryless nonlinearity
\begin {align*}
q(n) =&\; t(n) + (0.1+0.15i)\cdot t^2(n)\\
  &+ (0.06+0.05i)\cdot t^3(n).
\end{align*}
At the receiver end of the channel, the signal is corrupted by white Gaussian noise and then observed as $r(n)$.
The input signal that was fed to the channel had the form
\begin{align}\label{EQ:input}
s(n) = 0.70(\sqrt{1-\rho^2}X(n) + i\rho Y(n)),
\end{align}
where $X(n)$ and $Y(n)$ are gaussian random variables. This input is circular for $\rho=\sqrt{2}/2$ and highly non-circular if $\rho$ approaches 0 or 1 \cite{Adali10}. The aim of channel equalization is to construct an inverse filter which taking the output $r(n)$, reproduces the original input signal with as low an error rate as possible. To this end we apply the NCKLMS algorithm to the set of samples
\begin{align*}
\left((r(n+D), r(n+D-1), \dots, r(n+D-L))^T, s(n)\right),
\end{align*}
where $L>0$ is the filter length and $D$ the equalization time delay.

\begin{figure}
\begin{center}
\includegraphics[scale=0.44]{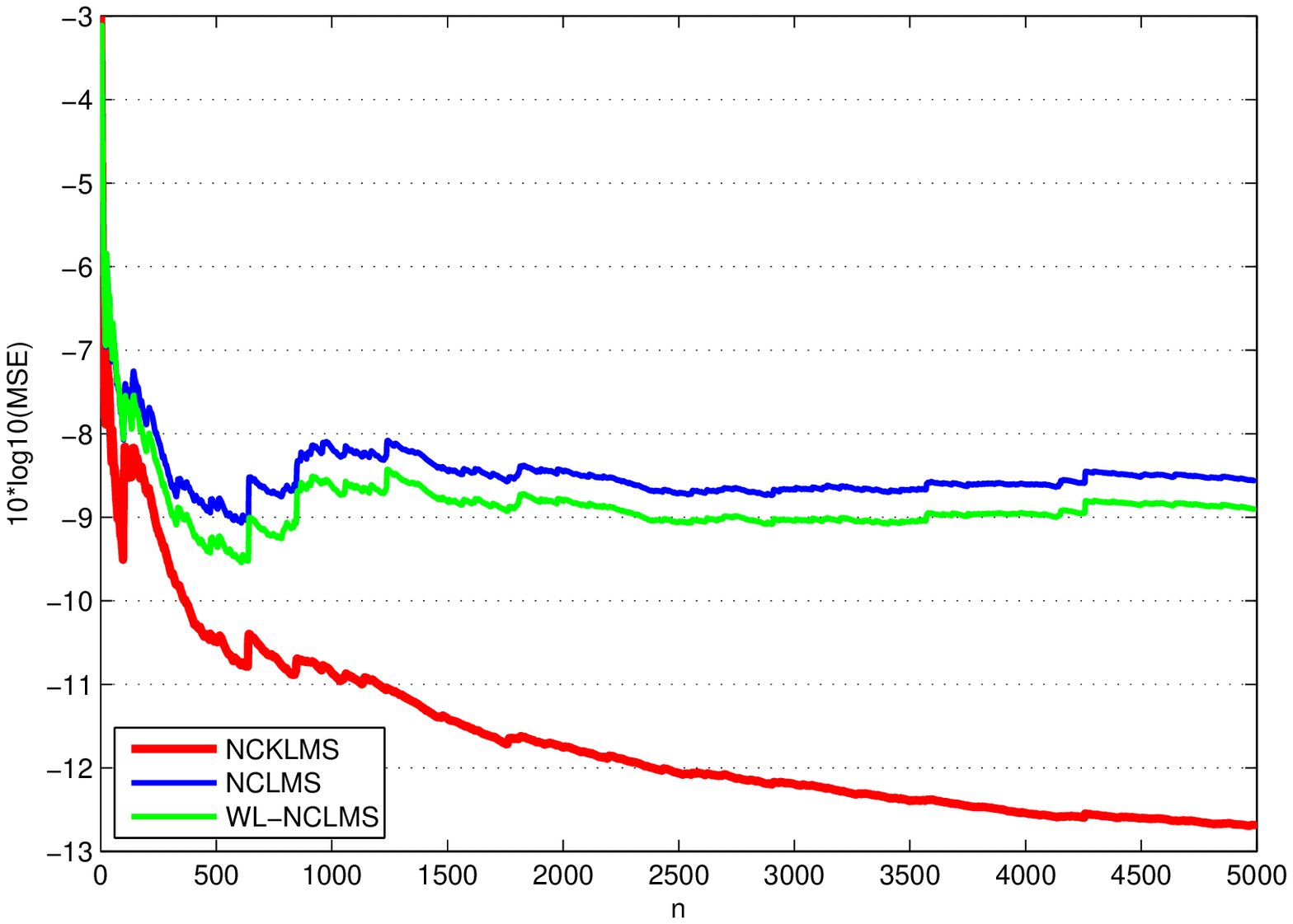}
\end{center}
\caption{Learning curves for KNCLMS ($\mu=1/2$), NCLMS ($\mu=1/16$) and WL-NCLMS ($\mu=1/16$) (filter length $L=5$, delay $D=2$) in the nonlinear channel equalization, for the non-circular input case ($\rho=0.1$).}\label{FIG:equal_noncircu}
\end{figure}

Experiments were conducted on a set of 5000 samples of the input signal (\ref{EQ:input}) considering both the circular and the non-circular case.
The results are compared with the NCLMS and the WL-NCLMS algorithms. In all algorithms the step update parameter $\mu$ is tuned for best possible results. Time delay $D$ was also set for optimality. Figures \ref{FIG:equal_circu} and \ref{FIG:equal_noncircu} show the learning curves of the NCKLMS using the Gaussian kernel $\kappa(\bx,\by)=\exp(-\|\bx-\by\|^2/\sigma^2)$ (with $\sigma=5$), compared with the NCLMS and the WL-NCLMS algorithms. Novelty criterion was applied to the NCKLMS for sparsification with $\delta_1=0.15$ and $\delta_2=0.2$. In both examples, NCKLMS considerably outperforms both the NCLMS and the WL-NCLMS algorithms. However, this enhanced behavior comes at a price in computational complexity, since the NCKLMS requires the evaluation  of the kernel function on a growing number of training examples.

\section{Conclusions}\label{SEC:Concl}
A new framework for kernel adaptive filtering for complex signal processing was developed. The proposed methodology employs a technique called complexification of RKHSs to construct complex RKHSs from real ones, providing the advantage of working with some popular real kernels in the complex domain. It has to be pointed out, that our method is a general one and can be used on any type of complex kernels that have or can been developed. To the best of our knowledge, this is the first time that a methodology for complex adaptive processing in RKHSs is proposed. Wirtinger's calculus has been extended to cope with the problem of differentiation in the involved (infinite) dimensional Hilbert spaces. The derived rules and properties of the extended Wirtinger's calculus on complex RKHS turn out to be similar in structure  to the special case of finite dimensional complex spaces. The proposed framework was applied on the complex LMS and the new complex Kernel LMS algorithm was developed. Experiments, which were performed on the equalization problem of a nonlinear channel for both circular and non-circular input data, showed a significant decrease in the steady state mean square error, compared with complex LMS and widely linear complex LMS.

\bibliographystyle{IEEEbib}
\bibliography{refs}

\end{document}